\documentclass{llncs}

\usepackage{times}
\usepackage{helvet}
\usepackage{courier}
\usepackage{slantsc}
\usepackage{tabu}
\usepackage{graphicx}
\usepackage{amsmath}
\usepackage{amssymb}
\usepackage{myamsthm}
\usepackage[algo2e,ruled,linesnumbered,vlined]{algorithm2e}

\theoremstyle{definition}

\newcommand{\K}{\hbox{\rm K}}
\newcommand{\M}{\hbox{\rm M}}
\newcommand{\ASPor}{\mbox{ {\tt or} }}
\newcommand{\ASPnot}{\mbox{ {\tt not} }}
\newcommand{\leftASPnot}{\mbox{{\tt not} }}

\newcommand{\thdrf}[1]{\footnotesize{\textsc{\textbf{#1}}}}

\newcommand{\equals}{\hspace{-0.07cm}=\hspace{-0.07cm}}

\newcommand{\AS}{\hbox{\rm AS}}
\newcommand{\Ep}[1]{\hbox{\ensuremath{{E\hspace{-0.08cm}_P}{(\hspace{-0.03cm}#1}\hspace{-0.03cm})}}}
\newcommand{\LEp}{\hbox{\ensuremath{\mathcal I_{\hspace{-0.02cm}E\hspace{-0.05cm}_P}}}}
\newcommand{\W}{\hbox{\ensuremath{W\hspace{-0.02cm}}}}

\newcommand{\card}[1]{\hbox{\ensuremath{\lvert\hspace{0.02cm}#1\hspace{0.01cm}\rvert}}}
\newcommand{\ASP}[1]{\hbox{\ensuremath{{\sf\scalebox{0.9}[0.9]{ASP}}(#1\hspace{-0.03cm})}}}

\newcommand{\APhi}{\hbox{\ensuremath{{\sf\scalebox{0.9}[0.9]{ASP}}(\Phi)}}}
\newcommand{\modKM}[1]{\hbox{\ensuremath{#1_{\tiny\setminus{\sf km}}}}}

\newlength\mylen

\def\k{\: \hbox{\rm K} \: }
\def\m{\: \hbox{\rm M} \: }
\def\true{\hbox{\sf\scalebox{0.9}[0.9]{True}}}
\def\false{\hbox{\sf\scalebox{0.9}[0.9]{False}}}
\def\AND{\hbox{\sf\scalebox{0.8}[0.9]{AND}} }
\def\enaf{\: \hbox{\bf not} \: }
\def\TM{\textsuperscript{\tiny\sf TM} }
\def\RTM{\textsuperscript{\textregistered} }

\title{{\fontsize{13pt}{18pt}\selectfont A Parallel Memory-efficient Epistemic
Logic Program Solver:\vspace{-0.05cm}\\Harder, Better, Faster}\vspace{-0.5cm}}
\author{{\sc Patrick Thor Kahl}\vspace{-0.05cm}\\
{\small Space and Naval Warfare Systems Center Atlantic, North Charleston, SC,
USA}\vspace{-0.05cm}\\
{\small{\tt patrick.kahl@navy.mil}}\vspace{0.125cm}\\
{\sc Anthony P. Leclerc}\vspace{-0.05cm}\\
{\small Space and Naval Warfare Systems Center Atlantic, North Charleston, SC,
USA}\vspace{-0.05cm}\\
{\small College of Charleston, Charleston, SC, USA}\vspace{-0.05cm}\\
{\small{\tt anthony.leclerc@navy.mil / leclerca@cofc.edu}}\vspace{0.125cm}\\
{\sc Tran Cao Son}\vspace{-0.42cm}}
\institute{New Mexico State University, Las Cruces, NM, USA\vspace{-0.05cm}\\
{\tt tson@cs.nmsu.edu}\vspace{-0.5cm}}

\begin{document}
\maketitle

\begin{abstract}
As the practical use of answer set programming (ASP) has grown with the
development of efficient solvers, we expect a growing interest in extensions of
ASP as their semantics stabilize and solvers supporting them mature. Epistemic
Specifications, which adds modal operators K and M to the language of ASP, is
one such extension. We call a program in this language an epistemic logic
program (ELP). Solvers have thus far been practical for only the simplest ELPs
due to exponential growth of the search space. We describe a solver that is
able to solve harder problems better (e.g., without exponentially-growing memory
needs w.r.t. K and M occurrences) and faster than any other known ELP solver.
\vspace{-0.3cm}
\end{abstract}

\begin{keywords}
Epistemic Logic Program Solver, Logic Programming Tools, Epistemic
Specifications, Answer Set Programming Extensions, Solver Algorithms
\vspace{-0.25cm}
\end{keywords}

\section{Introduction}

The language of \emph{Epistemic Specifications} \cite{Gelfond91a,Gelfond94a}
was introduced in the early 1990s by Michael Gelfond after observing the need
for more powerful introspective reasoning than that offered by answer set
programming (ASP) alone, extending ASP with modal operators $\K$ (``known'')
and $\M$ (``may be true'').
A program written in this language is called an \emph{epistemic logic program}
(ELP), with semantics defined using the notion of a \emph{world view}---a
collection of sets of literals (\emph{belief sets}), analogous to the answer
sets of an ASP program.
A renewed interest in Epistemic Specifications
\cite{Truszczynski11a,FaberWoltran11a} in 2011 included a proposed change to
the semantics by Gelfond \cite{Gelfond11a} in a preliminary effort to
avoid unintended world views. That work was continued by Kahl et al.
\cite{Kahl14a,KahlWatsonBalaiGelfondZhang15a} in the hopes of finding a
satisfactory semantics with respect to intuition and modeling of problems.
Later attempts to improve the semantics were offered by Su et al.
\cite{Su15a,FarinasHerzigSu15a} and more recently by Shen \& Eiter
\cite{ShenEiter16a} to further address unintended world view issues.

Along with maturation of the language were various attempts at developing a
solver or inference engine, including
{\tt ELMO} by Watson \cite{Watson94a},
{\tt sismodels} by Balduccini \cite{sismodels},
{\tt Wviews} by Kelly \cite{Wviews,Kelly07a} implementing Yan Zhang's algorithm
\cite{Zhang06a},
{\tt ESmodels} by Zhizheng Zhang et al.
\cite{ESmodels,CuiZhangZhao12a,ZhangZhaoCui13a,ZhangZhao14a},
and most recently {\tt ELPS} by Balai \cite{ELPS,BalaiKahl14a}.

In this paper, we present results of our efforts to implement a new ELP solver
that incorporates updated semantics and uses a scalable approach that
greatly decreases both the memory and time required for solving harder ELPs
compared to other solvers.

\section{Syntax and Semantics}\label{SyntaxAndSemantics}

In general, the syntax and semantics of the language of Epistemic Specifications
follow that of ASP with the notable addition of modal operators $\K$ and $\M$,
plus the new notion of a \emph{world view}. We assume familiarity with logic
programming---ASP in particular. For a good introduction, see any of
\cite{Baral03a,HandbookOfKR2008,GebserKaminskiKaufmannSchaub12a,GelfondKahl14a}.
For simplicity, the syntax presented does not include certain language features
(e.g., aggregates) of the proposed ASP core standard \cite{ASPCore2Standard},
but that does not mean such features should necessarily be considered excluded
from the language.
As previous authors have done before us, we will use $\AS(\mathcal{P})$ to
denote the set of all answer sets of an ASP program $\mathcal{P}.$

\subsection{Syntax}

An epistemic logic program is a set of rules of the form
$$\ell_1\ASPor ... \ASPor\ell_k\leftarrow g_1, ..., g_m,\ASPnot g_{m+1},
...,\ASPnot g_n.$$\noindent
where $k\ge 0$, $m\ge 0$, $n\ge m$, each $\ell_i$ is a literal (an atom or a
classically-negated atom), and each $g_i$ is either a literal (often called an
\emph{objective literal} within the context of Epistemic Specifications), or a
\emph{subjective literal} (a literal immediately preceded by $\K$ or $\M$).
As in ASP, a rule having a literal with a variable term is a shorthand for all
ground instantiations of the rule.

\subsection{Semantics}

It should be noted that the semantics described below differs somewhat from that
of \cite{Kahl14a,KahlWatsonBalaiGelfondZhang15a} as a world view has an
additional requirement based on \cite{ShenEiter16a}.

\begin{definition}{[When a Subjective Literal Is Satisfied]}\label{def1}\\
{\rm
Let $W$ be a non-empty set of consistent sets of ground literals, and
$\ell$ be a ground literal.
\begin{itemize}
\item \hspace{0.828cm}$\K\hspace{0.06cm}\ell\hspace{0.125cm}$ is satisfied by
$W$ if $~\forall\hspace{-0.01cm}A \in W:~\ell\in A$.
\item \ASPnot\hspace{0.02cm}$\K\hspace{0.06cm}\ell\hspace{0.125cm}$ is
satisfied by $W$ if $~\exists A \in W:~\ell\notin A$.
\item \hspace{0.808cm}$\M\hspace{0.04cm}\ell\hspace{0.02cm}$ ~is satisfied by
$W$ if $~\exists A \in W:~\ell\in A$.
\item \ASPnot$\M\hspace{0.04cm}\ell\hspace{0.02cm}$ ~is
satisfied by $W$ if $~\forall\hspace{-0.01cm}A \in W:~\ell\notin A$.
\end{itemize}
We will use symbol $\models$ to mean \emph{satisfies}; e.g.,
$W\models\K\hspace{0.05cm}\ell$ \hspace{0.02cm}means
\hspace{0.02cm}$\K\hspace{0.05cm}\ell$\hspace{0.02cm} is satisfied by $W.$
}
\end{definition}

\begin{definition}{[Modal Reduct]}\label{ModRedDef}\\
{\rm
Let $\Pi$ be a ground epistemic logic program and $W$ be a non-empty set of
consistent sets of ground literals.
We denote by $\Pi^W$ the \emph{modal reduct of $\Pi$ with respect to $W$}
defined as the ASP program\footnote{with nested expressions of the form
~$\leftASPnot\leftASPnot\ell$~ as defined in \cite{LifschitzTangTurner99a}}
obtained from $\Pi$ by replacing/removing subjective literals and/or deleting
associated rules in $\Pi$ per the following table:\\\noindent
\begin{tabular}{|l|l|l|}\hline
~{\small\bf subjective literal } $\varphi$~ &
~{\small\bf if~} $W\models\varphi$ {\small\bf ~then...}~ &
~{\small\bf if~} $W\not\models\varphi$ {\small\bf ~then...}~
\\\hline\hline
\hspace{1.65cm}$\K\hspace{0.06cm}\ell$ &
~replace ~$\K\hspace{0.06cm}\ell$~ with ~$\ell$ &
~delete rule containing ~$\K\hspace{0.06cm}\ell$ \\\hline
\hspace{0.82cm}\ASPnot\hspace{0.02cm}$\K\hspace{0.06cm}\ell$ &
~remove \ASPnot$\K\hspace{0.06cm}\ell$ &
~replace \ASPnot$\K\hspace{0.06cm}\ell$~ with ~$\ASPnot\ell$
\\\hline
\hspace{1.63cm}$\M\hspace{0.04cm}\ell$ &
~remove ~$\M\hspace{0.04cm}\ell$ &
~replace ~$\M\hspace{0.04cm}\ell$ with ~{\tt not}$\ASPnot\ell~$
\\\hline
\hspace{0.82cm}\ASPnot$\M\hspace{0.04cm}\ell$ &
~replace \ASPnot$\M\hspace{0.04cm}\ell~$ with $\ASPnot\ell~$ &
~delete rule containing \ASPnot$\M\hspace{0.04cm}\ell~$
\\\hline
\end{tabular}
}
\end{definition}

\newpage

\begin{definition}{[Epistemic Negations\footnote{introduced by Shen \&
Eiter in \cite{ShenEiter16a} using a different syntax (see the
appendix)}]}\label{EpDefinition}\\
{\rm
Let $\Pi$ be a ground epistemic logic program and $W$ be a non-empty set of
consistent sets of literals.
We denote by $\Ep{\Pi}$ the set of distinct subjective literals in $\Pi$ taking
the form \ASPnot$\K\hspace{0.04cm}\ell$\hspace{0.07cm} and 
\hspace{0.07cm}$\M\hspace{0.04cm}\ell$\hspace{0.05cm}
(called \emph{epistemic negations}) as follows:
$$
\Ep{\Pi}=\{\ASPnot\K\hspace{0.04cm}\ell :
\K\hspace{0.04cm}\ell\hspace{0.12cm}{\rm appears~in}~\Pi~\}
\cup
\{~\M\hspace{0.04cm}\ell :
\M\hspace{0.04cm}\ell\hspace{0.12cm}{\rm appears~in}~\Pi~\}.
$$
We use $\Phi$ to denote a subset of $\Ep{\Pi}$, and we denote by $\Phi_W$ the
subset of epistemic negations in $\Ep{\Pi}$ that are satisfied by $W;$ i.e.,
$~\Phi_W=\{~\varphi ~:~ \varphi\in\Ep{\Pi}~\land~W\models\varphi~\}.$
}
\end{definition}

\begin{definition}{[World View]}\label{WvDef}\\
{\rm
Let $\Pi$ be a ground epistemic logic program and $W$ be a non-empty set of
consistent sets of literals.\hspace{0.12cm}$W$ is a \emph{world view} of $\Pi$
if:
\begin{itemize}
\item $W\hspace{-0.1cm}=\hspace{-0.025cm}\AS(\Pi^W\hspace{-0.04cm})$; and
\item  there is no\hspace{0.13cm}$W^\prime$ such that\hspace{0.05cm}
$W^\prime\hspace{-0.1cm}=\hspace{-0.025cm}\AS(\Pi^{W^\prime}\hspace{-0.04cm})$
~and~ $\Phi_{W^\prime}\hspace{-0.02cm}\supset\Phi_W.$
\end{itemize}
\noindent
Note here the addition of a \emph{maximality requirement} on
$\Phi_W$\hspace{0.05cm}with respect to other guesses (corresponding to other
candidate world views) that is not in the semantics of
\cite{Kahl14a,KahlWatsonBalaiGelfondZhang15a}.
See \cite{ShenEiter16a} for discussion concerning the intuition behind the
proposed new semantics.
}
\end{definition}

\subsection{Discussion and Additional Definitions}

The semantics of an epistemic logic program $\Pi$ as described herein is
equivalent to the semantics described by Shen \& Eiter in \cite{ShenEiter16a}
for $\Pi$ translated to their syntax. The proof can be found in the appendix.

Although we prefer our syntax to that proposed in \cite{ShenEiter16a}, we find
the definition of an \emph{epistemic reduct} an excellent tool for describing
the new semantics, particularly with the emphasis on our ELP solver.
We therefore present the following additional definitions.

\begin{definition}{[Epistemic Reduct, Reduct-verifiable]}\label{EpRedDef}\\
{\rm
Let $\Pi$ be a ground epistemic logic program, $\Phi$ be a subset of
$\Ep{\Pi},$ and $\Psi{=}\Ep{\Pi} \setminus \Phi$. We denote by $\Pi^\Phi$ the
\emph{epistemic reduct of $\Pi$ with respect to $\Phi$} defined as the ASP
program obtained from $\Pi$ (assuming existence of corresponding $\W$) by
considering as
\begin{itemize}
\item[\bf\textsc{satisfied}:\hspace{0.55cm}~] \hspace{-0.7cm} the subjective
literals in $\Phi$ and the complements of the subjective literals in $\Psi$
\item[\bf\textsc{not satisfied}:\hspace{0.55cm}~]\hspace{-0.7cm} the subjective
literals in $\Psi$ and the complements of the subjective literals in $\Phi$%
\hspace{0.03cm}\footnote{$\K\hspace{0.05cm}\ell$ ~and
$\ASPnot\K\hspace{0.05cm}\ell$ ~are complements;~ $\M\hspace{0.05cm}\ell$ ~and
$\ASPnot\M\hspace{0.05cm}\ell$ ~are complements}
\end{itemize}
and taking actions according to the table given for the \emph{modal reduct} in
Definition \ref{ModRedDef}.
\medskip\\
For given $\Pi,$ we say that $\Phi$ is \emph{reduct-verifiable} if 
$W\hspace{-0.15cm}=\hspace{-0.075cm}\AS(\Pi^\Phi\hspace{-0.04cm})$, 
$W\hspace{-0.12cm}\neq\hspace{-0.07cm}\varnothing$, and 
$\Phi_W\hspace{-0.08cm}=\hspace{-0.07cm}\Phi.$
}
\end{definition}

\begin{definition}{[World View (alternative definition)]}\label{altWVdef}\\
{\rm
Let $\Pi$ be a ground ELP,
$\hspace{0.04cm}\Phi\hspace{-0.02cm}\subseteq\hspace{-0.02cm}\Ep{\Pi},$ and
$W\hspace{-0.15cm}=\hspace{-0.075cm}\AS(\Pi^\Phi\hspace{-0.04cm}).$
\hspace{0.04cm}$W$ is a \emph{world view} of $\Pi$ if:
\begin{itemize}
\item $\Phi$ ~is reduct-verifiable and
\item there exists no reduct-verifiable
$\Phi^\prime\hspace{-0.02cm}\subseteq\hspace{-0.02cm}\Ep{\Pi}$ such that
$\Phi^\prime\supset\Phi.$
\end{itemize}
}
\end{definition}

\section{Algorithms for Computing World Views}

Definitions \ref{EpRedDef} and \ref{altWVdef} show that a world
view of a program $\Pi$ can be computed by guessing a set
$\Phi\subseteq\Ep{\Pi}$ and verifying that $\Phi$ is maximal (with respect to
$\subseteq$) reduct-verifiable. As such, we use the term \emph{guess} to
refer to a set of elements from $\Ep{\Pi}$ in the descriptions of our
algorithms. For completeness of the paper, we include herein the basic
algorithm implemented in existing state-of-the-art ELP solvers
(Algorithm~\ref{algo:old}). A pictorial description of Algorithm~\ref{algo:old}
is given in Figure~\ref{AlgFigAB}(a). It is easy to see
Algorithm~\ref{algo:old} is complete if {\bf Translation} (Step 2) guarantees
all world views of $\Pi$ are disjoint subsets (groups) of the answer sets of
$\Pi^\prime.$ One such translation was proposed in
\cite{KahlWatsonBalaiGelfondZhang15a}.

\begin{algorithm2e}
 \DontPrintSemicolon
 \KwIn{$\Pi$ : an epistemic logic program}
 \KwOut{The set $\Omega = \{\omega : \omega\mbox{ is a world view of }\Pi\}$}

 $\Omega \gets \varnothing$\;

 {\bf Translation:} ASP program $\Pi^\prime$ is created from input ELP $\Pi$
 (effecting generation of $\Pi^\Phi$ for all guesses $\Phi\subseteq\Ep{\Pi}$).

 {\bf Computation:} $\Pi^\prime$ is solved using an ASP solver.

 {\bf Aggregation:} Answer sets of $\Pi^\prime$ are grouped according to
 corresponding $\Phi$.
  
 {\bf Verification:} Each group $G$ is verified (if $\Phi_G=\Phi$ then $G$ is
 added to $\Omega$). 
   
 \Return{$\Omega$}\;

 \caption{Compute World Views -- Old Algorithm}
 \label{algo:old}
\end{algorithm2e}

Although Algorithm~\ref{algo:old} is simple and easy to implement as an add-on
to an ASP solver, its main drawback lies in Steps 3 \& 4. Specifically,
it requires that \emph{all} answer sets of program $\Pi^\prime$ are computed
and then grouped before world views can be identified. This works well
for very small programs but can be impractical memory-wise since the number of
guesses grows exponentially with the number of subjective literals.
In addition, existing implementations of this algorithm do not allow for early
termination when the number of world views found reaches a user-specified limit
(e.g., 1). 

\subsection{New Algorithm} \label{CompOldNewSect}

\begin{figure}[h!]
\centering
\caption{Side-by-side Visual Comparison of the Old {\sf(a)} and New {\sf(b)}
         Algorithms}
\label{AlgFigAB}
\includegraphics[height=6cm,width=12cm,trim=.5cm 3cm .5cm -.5cm]{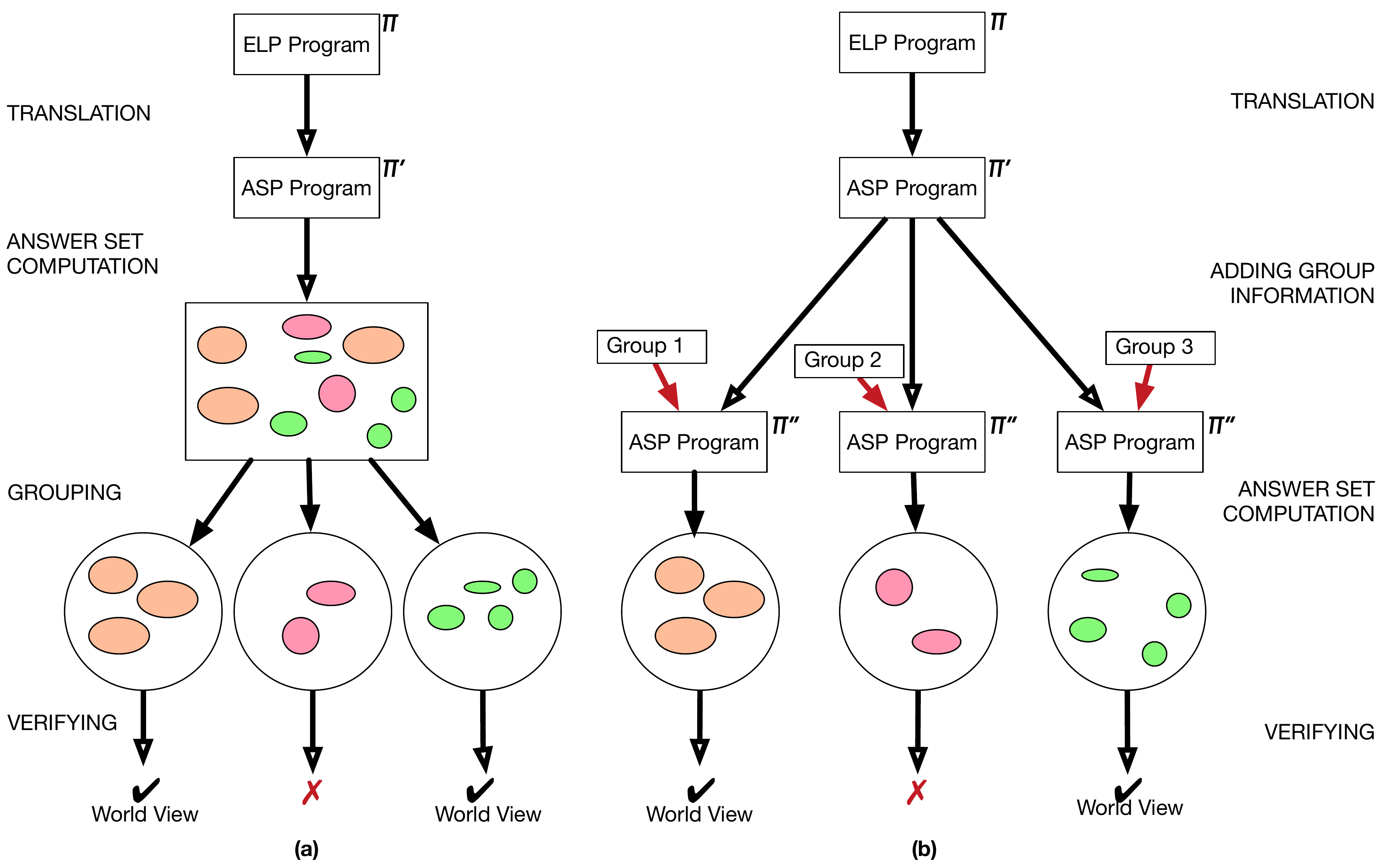}
\end{figure}

Our new algorithm improves Algorithm~\ref{algo:old} by addressing the memory
growth problem. It divides the work into parts instead of computing everything
in one call. The basic steps are given in Algorithm~\ref{algo:new}. A pictorial
description of the algorithm is given in Figure~\ref{AlgFigAB}(b).
In the form shown, the algorithm does not address the maximality requirement
of the updated semantics. Details of how the values of guesses are added to the
program in the {\bf Paritition} step (Step 4) will be described in the
implementation section, where selection of guesses is done systematically
via an {\em ordered search} with {\em pruning} that ensures the maximality
requirement is properly addressed. It is easy to see how limiting the number
of guesses during the {\bf Partition} step of Algorithm~\ref{algo:new} can
alleviate the memory concerns of Algorithm~\ref{algo:old}; however, as the
selection of guesses for each iteration is not specified, the algorithm can be
implemented in a variety of ways. The algorithm will thus be further refined in
Algorithms~\ref{algo:calc_wv_ord}, \ref{algo:calc_wv_group}, and
\ref{algo:calc_wv_par}.

\begin{algorithm2e}[h!]
 \DontPrintSemicolon
 \KwIn{$\Pi$ : an epistemic logic  program}
 \KwOut{The set $\Omega = \{\omega : \omega\mbox{ is a world view of }\Pi\}$}

 $\Omega \gets \varnothing$\;

 {\bf Translation:} ASP program $\Pi^\prime$ is created from input ELP $\Pi$
 (without the inclusion of guesses for subjective literals)

 \Repeat{all guesses tried}{

  {\bf Partition:} $\Pi^{\prime\prime}$ is created by adding ASP representation
  of guesses ($\Phi$ values) to $\Pi^\prime$

  {\bf Computation:} $\Pi^{\prime\prime}$ is solved using an ASP solver

  {\bf Aggregation:} Answer sets of $\Pi^{\prime\prime}$ are grouped according
  to corresponding $\Phi$.
  
  {\bf Verification:} Each group $G$ is verified (if $\Phi_G=\Phi$ then $G$ is
  added to $\Omega$). 
  
 }
 \Return{$\Omega$}\;

 \caption{Compute World Views -- New Algorithm}
 \label{algo:new}
\end{algorithm2e}

\vspace{-0.1cm}
Algorithm~\ref{algo:new} offers advantages over Algorithm~\ref{algo:old} by
providing a divide-and-conquer search method
that is configurable to allow for:
\begin{itemize}
 \item efficient use of memory,
 \item parallel processing, and
 \item early termination after a pre-specified number of world views found.
\end{itemize}

\vspace{-0.7cm}

\subsection{Instantiations of the New Algorithm: Overview}

The basic idea behind the new algorithm is to follow
Definition~\ref{altWVdef} in computing the world views. To ensure
completeness in an efficient manner, guesses need to be selected in a way that
facilitates satisfaction of the second condition of the definition. 
In our approach, guesses are ranked (into {\em levels})
in decreasing order by their cardinality. The cardinality of a guess,
$\lvert\Phi\rvert$, is called the {\em guess size}.
A guess is filtered out if it is a subset of a prior guess that was found
to be reduct-verifiable. Combined with the selection order, this eliminates
the possibility of a reduct-verifiable guess being a superset of any guess
that is not filtered out.
We therefore start with the guess $\Phi = \Ep{\Pi}$ and work systematically
down in order of guess size, filtering out guesses that are subsets of
previously found world views.
So to determine if a guess yields a world view, we need only compute the
answer sets of the epistemic reduct ($\Pi^\Phi$) and check that $\Phi$ is
reduct-verifiable. Algorithm~\ref{algo:calc_wv_ord} implements this idea. 
In this algorithm, $\APhi$ denotes the ASP representation of subjective
literals that are considered satisfied when creating
$\Pi^\Phi$\hspace{-0.07cm}, and $\Pi^{\prime}$ is constructed in such a way
that the corresponding $\AS(\Pi^{\prime\prime})$\footnote{modulo certain
fresh literals (see Section~\ref{AlgImpRep})}$=\AS(\Pi^\Phi).$
\vspace{-0.4cm}

\begin{algorithm2e}
  \DontPrintSemicolon
  \KwIn{$\Pi$ : an epistemic logic  program, $n_\Omega$: max number of world
  views desired}
  \KwOut{The set $\Omega = \{\omega : \omega\mbox{ is a world view of }\Pi\}$}

  $\Omega \gets \varnothing$\;

  $\Pi^{\prime} \leftarrow $  result of Step 2 of Algorithm~\ref{algo:new} \;

  \ForEach{subset $\Phi$ of $\Ep{\Pi}$ in decreasing order of $\card{\Phi}$}{
  
    \lIf(\tcp*[h]{{\hspace{-0.1cm}pruning filter}})
	 {$(\exists\omega\in\Omega : \Phi\subset\Phi_\omega)$}
	 {\bf continue~} \nllabel{line:filter}
      construct $\Pi^{\prime\prime} =
	             \Pi^{\prime} \cup \APhi$\; \nllabel{line:constr}
      compute $C = \AS(\Pi^{\prime\prime})$~
	   {\tt //\hspace{0.1cm}candidate world view}\; \nllabel{line:solve}
      \lIf(\tcp*[h]{{\hspace{-0.1cm}check if world view}})
	   {$(C \neq \varnothing) \land (\Phi_C=\Phi)$}{add $C$ to $\Omega$~}
      \lIf(\tcp*[h]{{\hspace{-0.1cm}return if desired \# world views
	   found\hspace{-0.1cm}}})
	   {$|\Omega| = n_\Omega$}{break ~}
  }
  \Return{$\Omega$}\;

  \caption{Compute World Views -- Level Order, One Guess at a Time}
  \label{algo:calc_wv_ord}
\end{algorithm2e}

\vspace{-0.4cm}
Note the pruning filter on Line~\ref{line:filter} of
Algorithm~\ref{algo:calc_wv_ord} that, combined with the search order,
implements the maximality requirement of Definition~\ref{altWVdef}. 
This filter can be turned off to revert to the semantics of
\cite{KahlWatsonBalaiGelfondZhang15a}.
We refer to this filter later in
Algorithms~\ref{algo:calc_wv_group}~\&~\ref{algo:calc_wv_par}
without explicit pseudocode when we say {\em filtered}.

\begin{figure}[t]
\centering
\caption{Visual Depiction of Algorithm~\ref{algo:calc_wv_group}}
\label{AlgFigC}
\includegraphics[height=8.5cm,width=9cm,trim=.2cm .2cm .4cm -.5cm]{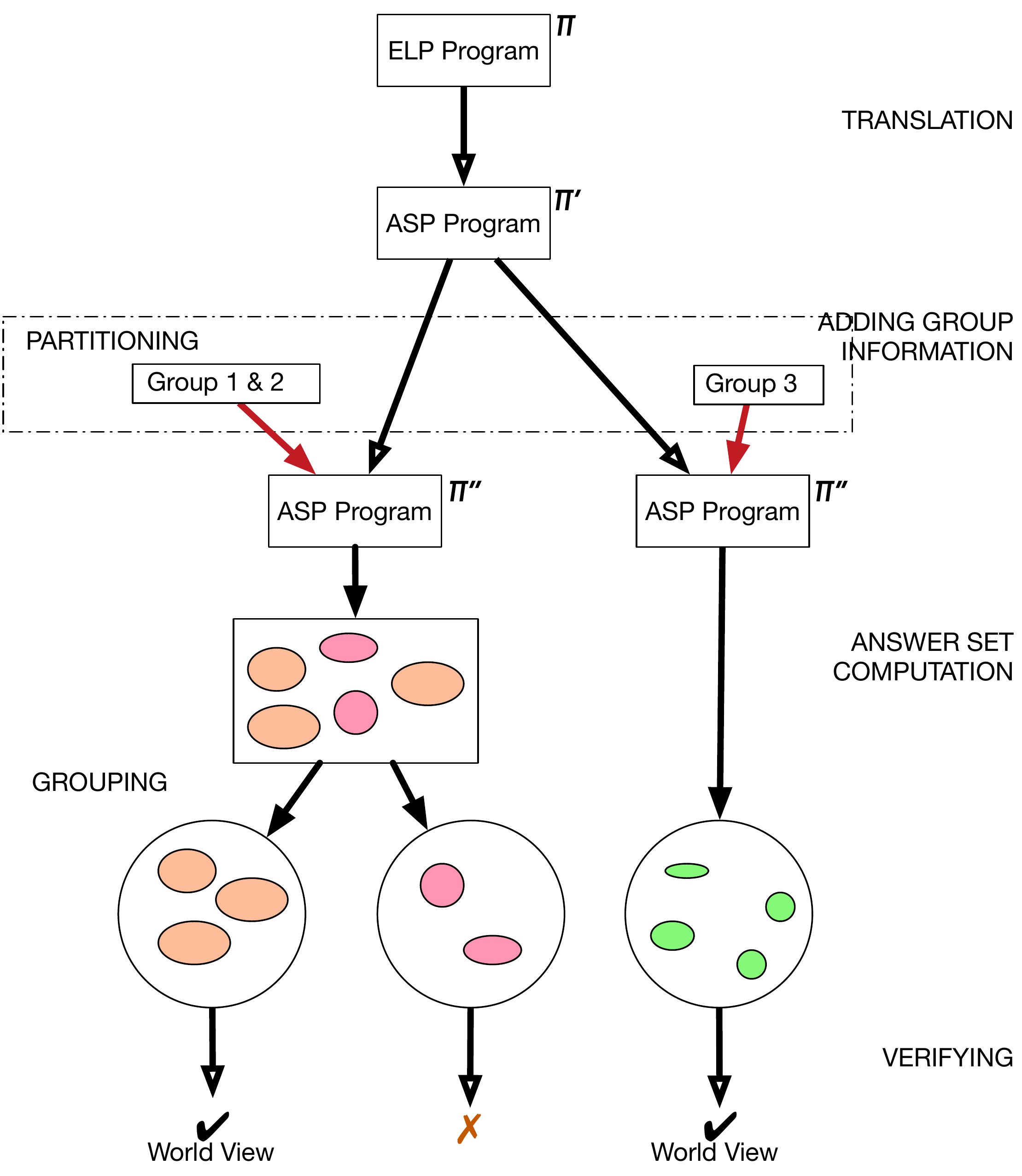}
\end{figure}

Although Algorithm~\ref{algo:calc_wv_ord} solves the memory problem, it
increases the number of calls to the ASP solver. As such, the algorithm can be
inefficient. To address this concern, we
implement Algorithm~\ref{algo:calc_wv_group} that computes the world views for
multiple guesses at the same time. The trade-off is that grouping (the
{\bf Aggregation} step of Algorithm~\ref{algo:new}) must be performed.
Additionally, care needs to be taken to ensure completeness and avoid
repetition. We accomplish this by requiring that
all guesses in a set have the same guess size,
all sets of guesses are pairwise disjoint, and
all guesses of the same size are tried (or filtered out) before moving to
the next level.
This is achieved as follows. 
We introduce parameter $n_G$ representing the (maximum) number of
guesses per ASP solver call, and partition each level
$L_k = \{ \Phi : (\Phi \subseteq \Ep{\Pi}) \land (\card{\Phi} = k) \}$
(for $0 \leq k \leq \lvert\Ep{\Pi}\rvert$)
into groups of at most $n_G$ guesses.
Detail on the implementation is given in the next subsection.
Figure~\ref{AlgFigC} shows a visual description of this algorithm.
 
Finally, in order to utilize the computing power of modern computers and
enhance the performance of the solver, we implement a parallel version of
the algorithm.
For brevity, we omit parameter $n_\Omega$ and associated pseudocode for early
termination when desired number of world views is found.

\SetKw{KwDownTo}{downto}
\begin{algorithm2e}
  \DontPrintSemicolon
  \KwIn{$\Pi$ : an epistemic logic  program,
        $n_G$ : \#groups per ASP solver call
  }
  \KwOut{The set $\Omega = \{\omega : \omega\mbox{ is a world view of }\Pi\}$}

  $\Omega \gets \varnothing$\;

  $\Pi^{\prime} \leftarrow $  result of Step 2 of Algorithm~\ref{algo:new} \;
  
  \For(\tcp*[h]{{\hspace{-0.1cm}for each level do $\ldots$}})
   {$k \leftarrow \card{\Ep{\Pi}}$ \KwDownTo $0$}{
  
    Partition $L_k$ into $t = \left\lceil \frac{\card{L_k}}{n_G} \right\rceil$
	groups $G^1_k,\ldots,G^t_{k}$ such that $|G^i_k| \le  n_G$
     
    \ForEach{filtered group $G^i_k$ of $L_k$}{
      construct $\Pi^{\prime\prime} = \Pi^{\prime} \cup \mathbf{ASP}(G^i_k)$ \;
      compute $G^{\mathit{as}} = \AS(\Pi^{\prime\prime})$\;
      \ForEach(\tcp*[h]{{\hspace{-0.1cm}group}})
	   {associated candidate world view ~$C$~ of ~$G^{\mathit{as}}$~}{
        \lIf(\tcp*[h]{{\hspace{-0.1cm}check if world view}})
		 {$(C \neq \varnothing) \land (\Phi_C=\Phi)$}{add $C$ to $\Omega$~}
      }
    }
  }
  \Return{$\Omega$}\;

  \caption{Compute World Views -- Level Order, Multiple Guesses at a Time}
  \label{algo:calc_wv_group}
\end{algorithm2e}

\SetKw{KwDownTo}{downto}
\SetKwFor{ForPar}{for}{do in parallel}{endfpar}
\begin{algorithm2e}
  \DontPrintSemicolon
  
  \KwIn{$\Pi$ : an epistemic logic  program,
        $n_G$ : \#groups per ASP solver call,
		$n_p$~:~\#processes
  }
  \KwOut{The set $\Omega = \{\omega : \omega\mbox{ is a world view of }\Pi\}$}

  $\Omega \gets \varnothing$\;

  $\Pi^{\prime} \leftarrow $  result of Step 2 of Algorithm~\ref{algo:new} \;
  
  \For(\tcp*[h]{{\hspace{-0.1cm}for each level do $\ldots$}})
   {$k \leftarrow \card{\Ep{\Pi}}$ \KwDownTo $0$}{

    Partition $L_k$ into $t = \left\lceil \frac{\card{L_k}}{n_G}
	\right\rceil$ groups $G^1_k,\ldots,G^t_{k}$  such that
	$|G^i_k| \le  n_G$\; \nllabel{line:partition}
    Mark every group as {\em not considered} 

          \Repeat{every group is either marked as considered or filtered out}{
      \For(\tcp*[h]{{\hspace{-0.1cm}for each processor do $\ldots$}})
	   {$i \leftarrow 1$ \KwTo $n_p$}{
        select a {\em filtered group} $G$ from $G^1_k,\ldots,G^t_{k}$ that is
		 marked as {\em not considered}\;
		mark group $G$ as {\em considered}\;   
        construct $\Pi^{\prime\prime}_i = \Pi^{\prime}\cup\mathbf{ASP}(G)$  \;
      }
      \ForPar(\tcp*[h]{{\hspace{-0.1cm}solve in parallel}})
	   {$i \leftarrow 1$ \KwTo $n_p$}{
        compute $G^{\mathit{as}}_i = \AS(\Pi^{\prime\prime}_i)$\;
      }
	  \tcp*[h]{accumulate the results of each processor and $\ldots$}\;
      \For{$i \leftarrow 1$ \KwTo $n_p$}{
        \ForEach(\tcp*[h]{{\hspace{-0.1cm}group}}){associated candidate world
		 view ~$C$~ of ~$G^{\mathit{as}}_i$~}{
          \lIf(\tcp*[h]{{\hspace{-0.1cm}check\hspace{0.125cm}if\hspace{0.025cm}
		   world\hspace{0.125cm}view\hspace{-0.15cm}}})
		   {$(C\neq\varnothing) \land (\Phi_C=\Phi)$}{add $C$ to $\Omega$}
        }
      }
    }
  }
  \Return{$\Omega$}\;

  \caption{Compute World Views -- Parallel Version}
  \label{algo:calc_wv_par}
\end{algorithm2e}

\subsection{Implementation Representation}\label{AlgImpRep}

Let $\Pi$ be a ground ELP and $\Ep{\Pi}$ be its set of epistemic negations.
Let $\LEp$ be an enumeration of $\Ep{\Pi}$ and for each $\varphi\in\Ep{\Pi}$,
let $ord(\varphi)$ be the index of $\varphi$ in $\LEp$. 
We represent each guess $\Phi\subseteq\Ep{\Pi}$ with a bitvector of length
$n=\lvert\Ep{\Pi}\rvert$ as follows:
$B_\Phi = \left[ b_1, b_2, ..., b_n \right]$
where for $i = ord(\varphi)$, bit $b_i = 1$ iff $\varphi\in\Phi.$ 

Each bitvector representing a guess can be represented by an integer $X$ such
that $0 \le X \le 2^n{-}1$; thus, each $X$ represents a unique subset of
$\Ep{\Pi}$ and vice versa.
This one-to-one correspondence is useful in the implementation
as it allows for fast checking of the subset relation between guesses (Line 4,
Algorithm~\ref{algo:calc_wv_ord}).
For later use, we denote by ~$\operatorname{popcount}(X)$~ the number of one
bits in the binary representation of $X$. 

As an example, consider $\Ep{\Pi} = \{ \leftASPnot
\K\hspace{0.06cm}\ell_{1}, \ASPnot \K\hspace{0.06cm}\ell_{2},~
\M\hspace{0.06cm}\ell_{3} \}$.
Assuming we enumerate the elements of $\Ep{\Pi}$ from left to right, the
bitvector $B_\Phi = \left[ b_1, b_2, b_3 \right]$ would be constructed with
bit $b_1$ representing the truth value
(w.r.t. membership in $\Phi$, where $1{=}\true$ and $0{=}\false$) of
$\leftASPnot \K\hspace{0.06cm}\ell_{1}$,
$b_2$ representing the truth value of
$\leftASPnot \K\hspace{0.06cm}\ell_{2}$, and $b_3$ representing the truth
value of
$\M\hspace{0.06cm}\ell_{3}$. For example, a bitvector $B_\Phi
= \left[ 0, 1, 1\right]$ would represent $\Phi = \{
\leftASPnot \K\hspace{0.06cm}\ell_{2},~ \M\hspace{0.06cm}\ell_{3} \}$.

For the translation of an ELP $\Pi$ to an ASP program, we use a variant of
the one developed in \cite{KahlWatsonBalaiGelfondZhang15a}. For each 
literal $\ell$ in the epistemic negations of $\Ep{\Pi}$ of the form
$\leftASPnot\K\hspace{0.05cm}\ell$,
let $k\_\ell$, $k0\_\ell$, and $k1\_\ell$ be fresh atoms
created by prefixing $\ell$ with $k\_$, $k0\_$, and $k1\_$ (respectively),
substituting $2$ for $\neg$ if $\ell$ is a classically-negated atom.
Likewise, for epistemic negations of the form $\M\hspace{0.05cm}\ell$ in
$\Ep{\Pi}$,
let $m\_\ell$, $m0\_\ell$, $m1\_\ell$ be fresh atoms created in like fashion.
For example, if $\ell\equals p(a)$ then $k\_\ell$ denotes $k\_p(a)$, but if
$\ell\equals\neg p(a)$ then $k\_\ell$ denotes $k\_2p(a)$. An atom denoted by
$k\_\ell$, $k0\_\ell$, or $k1\_\ell$ will be referred to as a \emph{k-atom},
whereas an atom denoted by $m\_\ell$, $m0\_\ell$, or $m1\_\ell$ will be
referred to as an \emph{m-atom}. By a \emph{k-/m-literal} we mean a k-/m-atom
or its negation. For a set of sets of literals $C$, we use the notation
$\modKM{C}$ to mean $C$ \emph{modulo k-/m-literals} (i.e., with k-/m-literals
removed from the sets of $C$).

Intuitive meanings of the atoms are:
$k1\_\ell$ stands for ``$\K\hspace{0.08cm}\ell$ is \true,''
$k0\_\ell$ stands for ``$\K\hspace{0.08cm}\ell$ is \false,''
$m1\_\ell$ stands for ``$\M\hspace{0.06cm}\ell$ is \true,''
and $m0\_\ell$ stands for ``$\M\hspace{0.06cm}\ell$ is \false.''
We can thus view $k1\_\ell$ as corresponding to $\K\hspace{0.05cm}\ell,$
\hspace{0.02cm}$k0\_\ell$ as corresponding to
$\leftASPnot\K\hspace{0.05cm}\ell,$ etc. 

\subsection{Implementation Details}

We describe here implementation details for Algorithm~\ref{algo:calc_wv_par}.
\begin{enumerate}

\item\label{algstep1} {\bf Translation}: Given ground ELP $\Pi$, we create
  ASP program $\Pi^{\prime}$ by
  ({\em i}) leaving rules without subjective literals unchanged; and
  ({\em ii}) for rules containing subjective literals, replacing
  subjective literals and adding new rules per the following table.
  \begin{center} 
  \begin{tabular}{|m{1.85cm}|m{2.20cm}|m{4.20cm}|}\hline
  {\tiny~}\thdrf{subjective}\vspace{-0.1cm}\newline\vspace{-0.1cm}\thdrf{$\phantom~~~~$literal}&
  \thdrf{~~~~replace}\vspace{-0.1cm}\newline\vspace{-0.1cm}\thdrf{$\phantom~~~~~~~$with}&
  ~\thdrf{add} ~\vspace{-0.1cm}\newline\vspace{-0.1cm}
  \thdrf{rules} ~\\\hline\hline
  \vspace{ 0.10cm}\hspace{1.00cm}$\K\hspace{0.08cm}\ell$                          &
  \vspace{ 0.10cm}$~\ASPnot\neg k\_\ell,~\ell$                                    &
  \vspace{ 0.15cm}$~\neg k\_\ell\hspace{0.08cm}\leftarrow k0\_\ell.$              \\\cline{1-2}
  \vspace{ 0.10cm}\hspace{0.20cm}$\ASPnot\K\hspace{0.08cm}\ell$                   &
  \vspace{ 0.10cm}\hspace{0.90cm}$\neg k\_\ell$                                   &
  \vspace{-0.05cm}$~\neg k\_\ell\hspace{0.08cm}\leftarrow k1\_\ell,~\ASPnot\ell.$ \\\hline
  \vspace{ 0.10cm}\hspace{1.00cm}$\M\hspace{0.05cm}\ell$                          &
  \vspace{ 0.10cm}\hspace{0.98cm}$m\_\ell$                                        &
  \vspace{ 0.15cm}$~m\_\ell\leftarrow m1\_\ell.$                                  \\\cline{1-2}
  \vspace{ 0.10cm}\hspace{0.20cm}$\ASPnot\M\hspace{0.05cm}\ell$                   &
  \vspace{ 0.10cm}\hspace{0.15cm}$\ASPnot\hspace{0.02cm}m\_\ell$                  &
  \vspace{-0.05cm}$~m\_\ell\leftarrow m0\_\ell,~\ASPnot\leftASPnot\ell.$          \\\hline
  \end{tabular}
  \end{center}

  We note that this translation is slightly different from the translation in
  \cite{KahlWatsonBalaiGelfondZhang15a} in that it does not add the rules for
  guessing the values of elements in $\Ep{\Pi}$. On the other hand, it implies
  that if equivalent rules for guessing the values of elements in $\Ep{\Pi}$
  are added to $\Pi^\prime$ then correctness of the algorithm is maintained.
  This property is guaranteed by the partitioning step of the algorithm.
  \smallskip

\item {\bf Partition:} The algorithm employs a bitvector representation for
  guesses in partitioning the search space. The partitioning of a level,
  $L_k$, occurs on Line~\ref{line:partition} of the algorithm.
  Each group of bitvectors, $G^1_k,\ldots,G^t_{k}$ for
  $t = \left\lceil \frac{\card{L_k}}{n_G} \right\rceil$\hspace{-0.03cm},
  is produced ``on demand'' with group size at most $n_G,$ and $G^1_k$
  containing the first $n_G$ elements of
  $L_k,$ $G^2_k$ containing the next $n_G$ elements of $L_k,$ etc.
  Partitioning is accomplished using a generating function ``seeded'' with an
  appropriate bitvector of length $n$ with $k$ one bits.  Each subsequent call
  to the generating function produces the ``next'' bitvector of length $n$ with
  $k$ one bits.  Thus, storage is required only for representing groups of
  guesses currently under consideration.
  \smallskip

\item $\Pi^{\prime\prime} = \Pi^\prime \cup \ASP{G}$, where $G$ is a set of
  bitvectors, is implemented as
  $\Pi^\prime \cup \ASP{\{X_B : (B \in G) \wedge (X_B \:
  \textnormal{is the integer whose binary representation is}\: B) \}}$.
  \smallskip

\item\label{algstep5c}\label{alggroup} {\bf Aggregation:}
  Answer sets of $\Pi^{\prime\prime}$ are grouped 
  by common k-/m-atoms of the form $k0\_\ell$, $k1\_\ell$,
  $m0\_\ell$, and $m1\_\ell$, each group representing a candidate world view.
  It is easy to see that a group's k-/m-atoms correspond to a guess
  $\Phi_X$ for some $X \in G$.
  \smallskip
 
\item\label{algstep5d}\label{algverify} {\bf Verification:} For each group $C$ 
    representing a candidate world view, check that the following conditions
	are met for all its k-/m-atoms:
	\medskip

    $~~~~$\begin{minipage}{11.3cm}
      \begin{enumerate}
      \item if \hspace{0.13cm}$k1\_\ell$\hspace{0.09cm} is in the sets
        of $C$, then $\ell$ is in every set of $C$;
      \item if \hspace{0.13cm}$k0\_\ell$\hspace{0.09cm} is in the sets
        of $C$, then $\ell$ is missing from at least one set of $C$;
      \item if ~$m1\_\ell$\hspace{0.02cm} is in the sets of $C$, then
        $\ell$ is in at least one set of $C$; and
      \item if ~$m0\_\ell$\hspace{0.02cm} is in the sets of $C$, then
        $\ell$ is missing from every set of $C$.
      \end{enumerate}
    \end{minipage}\medskip
	
    $\modKM{C}$ is a \emph{world view} of $\Pi$ if the conditions above are
	met.

\end{enumerate}

\vspace{-0.3cm}
\subsection{Correctness of Algorithm}
The correctness of the algorithm follows from the proofs for soundness
and completeness given in \cite{KahlWatsonBalaiGelfondZhang15a} when one
considers that we are simply partitioning the search space into groups of
manageable size (rather than generating all possible combinations of
subjective literal truth values en masse) and imposing a popcount-level
search order. It is easy to see that the maximality requirement of the new
semantics is guaranteed by the level-based search order and filtering by
considering that:
\vspace{-0.1cm}
\begin{itemize}
\item if a set of answer sets $C$ corresponding to $\Phi_C$ is found that
satisfies the verification conditions of Step \ref{algverify}, then the ordered
search and the filtering out of any corresponding $\Phi\subset\Phi_C$ ensures
that no previous world view $W$ was found corresponding to $\Phi_W$ such that
$\Phi_W\supset\Phi_C;$ ~and
\item if a set of answer sets $C$ corresponding to $\Phi_C$ is found to be
a world view, then any set $C^\prime$ corresponding to $\Phi_{C^\prime}$
where $\Phi_{C^\prime}\subset\Phi_C$ will be filtered out thereafter.
\end{itemize}
\vspace{-0.3cm}

\section{Test Results}

We tested an implementation of our algorithm on a Dell\TM Precision\TM T3500
server with an Intel\RTM Xeon\RTM W3670@3.2GHz with 6 cores
and 12 GB RAM running CentOS v6.7 operating system.
For the front-end, we used the {\tt ELPS} solver
by Balai \cite{ELPS} to convert an ELPS program (see \cite{BalaiKahl14a}) to an
associated ASP program, grounding it with ASP grounder {\tt gringo} by Kaminski
\cite{Potassco} in order to obtain what would be the result of Step 2
of our algorithm for a corresponding ground ELP from the
ungrounded ELPS program. We then took the resulting ground ASP program as input
to our solver, {\tt ELPsolve}---a loosely-coupled system that uses ASP solver
{\tt clingo} by Kaminski \& Kaufmann \cite{Potassco}.

This method allowed us to directly compare our solver performance with that of
{\tt ELPS} for the same input programs. The following table shows our results,
giving the best run time (elapsed time in seconds) observed over the course of
testing. A dash (--) in the {\tt ELPS} column indicates that a runtime
error occurred due to insufficient memory.\smallskip\\
\noindent
\begin{tabular}{|c|r|r|r||c|r|r|r|}\hline\vspace{-0.01cm}
~{\footnotesize\textbf{$\Pi$}}~             &
~{\footnotesize\textbf{$\card{\Ep{\Pi}}$}}~ &
~{\footnotesize\textbf{\tt ELPS}}~          &
~{\footnotesize\textbf{\tt ELPsolve}}~      &
~{\footnotesize\textbf{$\Pi$}}~             &
~{\footnotesize\textbf{$\card{\Ep{\Pi}}$}}~ &
~{\footnotesize\textbf{\tt ELPS}}~          &
~{\footnotesize\textbf{\tt ELPsolve}}~      \vspace{-0.03cm}\\\hline\vspace{-0.04cm}
~{\footnotesize\tt eligible01}~ & ~2~  & ~$<$1s~ & ~$<$1s~  &
~{\footnotesize\tt yale1}~      & ~5~  & ~$<$1s~ & ~$<$1s $~$ \\\hline
~{\footnotesize\tt eligible06}~ & ~12~ & ~1s~    & ~$<$1s~  & 
~{\footnotesize\tt yale2}~      & ~8~  & ~$<$1s~ & ~$<$1s $~$ \\\hline
~{\footnotesize\tt eligible08}~ & ~16~ & ~16s~   & ~1s~     & 
~{\footnotesize\tt yale3}~      & ~11~ & ~$<$1s~ & ~$<$1s $~$ \\\hline
~{\footnotesize\tt eligible09}~ & ~18~ & ~150s~  & ~2s~     & 
~{\footnotesize\tt yale4}~      & ~14~ & ~$<$1s~ & ~$<$1s $~$ \\\hline
~{\footnotesize\tt eligible10}~ & ~20~ & ~--~    & ~4s~     & 
~{\footnotesize\tt yale5}~      & ~17~ & ~13s~   & ~$<$1s $~$ \\\hline
~{\footnotesize\tt eligible12}~ & ~24~ & ~--~    & ~119s~   & 
~{\footnotesize\tt yale6}~      & ~26~ & ~--~    & ~$<$1s $~$ \\\hline
~{\footnotesize\tt eligible14}~ & ~28~ & ~--~    & ~1667s~  & 
~{\footnotesize\tt yale7}~      & ~30~ & ~--~    & ~2s $~$    \\\hline
~{\footnotesize\tt eligible16}~ & ~32~ & ~--~    & ~16124s~ & 
~{\footnotesize\tt yale8}~      & ~34~ & ~--~    & ~76s   $~$ \\\hline
\end{tabular}\smallskip

The {\tt eligible}$N$ programs are ELPS implementations of the
\emph{scholarship eligibility problem} (see \cite{Gelfond91a}) with $N$
corresponding to the number of students.
The {\tt yale}$N$ programs are ELPS implementations of a version of the
\emph{Yale shooting problem} (see \cite{HanksMcDermott87a}) that use the
\emph{epistemic conformant planning module} defined in
\cite{KahlWatsonBalaiGelfondZhang15a} with $N$ corresponding to the horizon
(number of steps in a plan). For the {\tt yale}$N$ programs, {\tt ELPsolve}
incorporates two heuristics found during analysis of the conformant planning
module:
\begin{enumerate}
\item reduction of the search space size by a factor of 4 by recognizing that
two specific subjective literals must be satisfied in a world
view;\footnote{This heuristic is generalizable, though a thorough discussion
will be left for a future paper.} and
\item the horizon corresponds to a particular level (guess size) for
searching.
\end{enumerate}

\begin{figure}[h!]
\centering
\caption{Run Time vs. \#Processors (for program {\tt eligible12})}
\label{ProcTimesFig}
\includegraphics[height=9cm,width=12cm,trim=1.5cm 13.0cm 1.5cm 1.0cm]{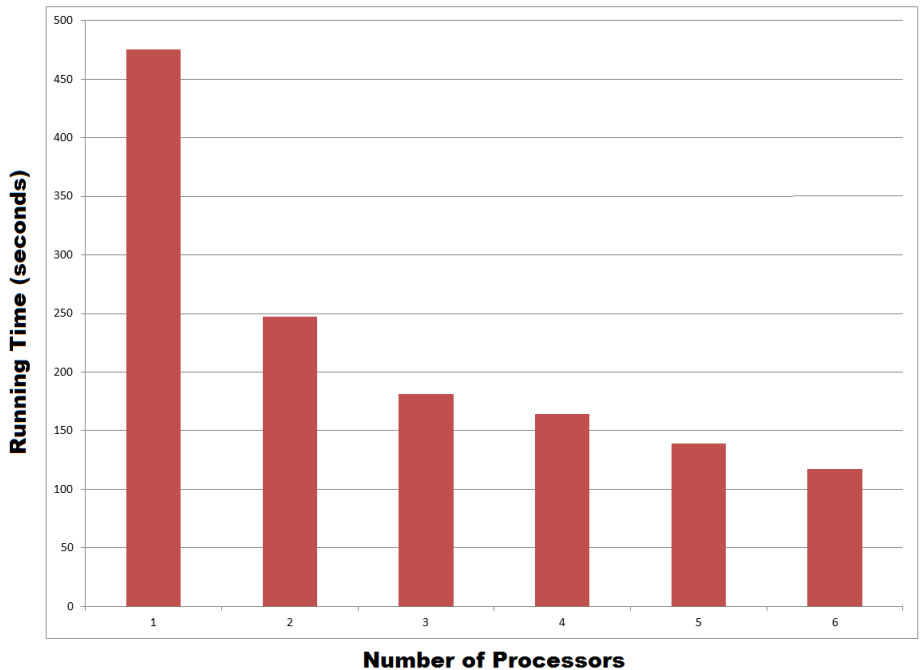}
\end{figure}

Figure~\ref{ProcTimesFig} shows how run times for solving {\tt eligible12}
with {\tt ELPsolve} improve with an increase in the number of processors used.
Note that although the 6-core Xeon processor in our test machine is
hyperthread-enabled, it is unlikely that there would be any significant
speed-up by simply increasing the number of processors assigned to multi-task
beyond 6 as our implementation uses 2-thread multi-threading for each task.

\vspace{-0.5cm}

\begin{figure}[h!]
\centering
\caption{Memory Usage (MB): {\tt ELPS} vs. {\tt ELPsolve}}
\label{MemUseFig}
\includegraphics[height=8.5cm,width=12cm,trim=0.2cm 12.0cm 0.3cm 0cm]{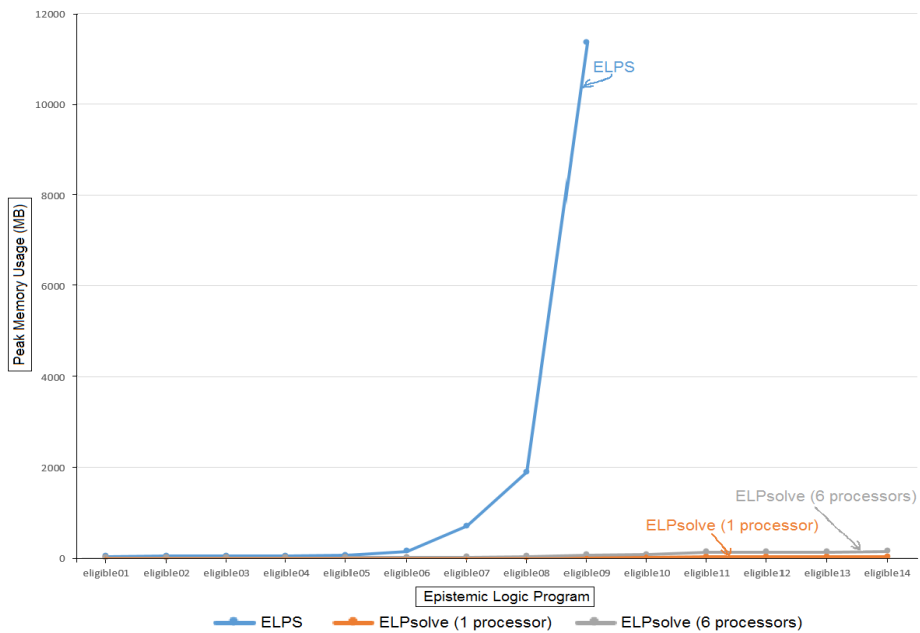}
\end{figure}

\vspace{-0.2cm}
Figure~\ref{MemUseFig} shows the peak memory usage for {\tt ELPS} and
{\tt ELPsolve} (max \#guesses per ASP call set at 300 for both 1 and 6
processors) when solving various {\tt eligible}$N$ programs. Note the blue
line curving sharply upward for {\tt ELPS} as $N$ grows whereas the
memory required for {\tt ELPsolve} remains relatively low and flat.

\section{Related Work}

\vspace{-0.2cm}
In \cite{Zhang06a}, Yan Zhang investigated computational properties of
epistemic logic programs, leading to the development of an algorithm for
computing world views. Michael Kelly implemented the algorithm as solver
{\tt Wviews} \cite{Wviews}, a project for his Honours Thesis.

Cui, Zhizheng Zhang and Zhao \cite{CuiZhangZhao12a} investigated the problem of
grounding an epistemic logic program. That work culminated in the development
of a grounder known as {\tt ESParser}. Their more recent efforts
\cite{ZhangZhaoCui13a,ZhangZhao14a} include the development of an associated
solver called {\tt ESmodels} \cite{ESmodels}. See
\cite{KahlWatsonBalaiGelfondZhang15a} for a comparison of solver performance
between {\tt ESmodels} and {\tt ELPS}.

Balai and Kahl \cite{BalaiKahl14a} extended Epistemic Specifications by adding
a sorted signature. A modification of the algorithm in
\cite{KahlWatsonBalaiGelfondZhang15a,Kahl14a} was implemented by Balai to
develop solver {\tt ELPS} \cite{ELPS}. Our solver complements this work by
addressing the memory growth problem and updating it for the new semantics.

Zhizheng Zhang and Shutao Zhang \cite{ZhangZhang15a} investigated combining
ideas from Graded Modal Logic with ASP. Their langauge can be
viewed as an extension of Epistemic Specifications, adding the modal concepts
``at least as many as'' and ``at most as many as'' to the language. They
continued this work with Wang \cite{ZhangWangZhang15a}, referring to the
language as GI-log. A GI-log solver called {\tt GIsolver} \cite{ESmodels} was
developed using a generate-and-test algorithm similar (at a high level) to the
one described in Algorithm~1.

Shen and Eiter \cite{ShenEiter16a} proposed the updated semantics used in this
paper and implemented in our solver, albeit using different syntactic notation.
Their work resolved most of the open questions raised in \cite{Kahl14a} and
provided inspiration for our algorithm.

\section{Conclusions and Future Work}

We improved the algorithm and developed a solver for epistemic logic programs
that:
\begin{itemize}
\item incorporates the latest semantics (as proposed by Shen \& Eiter
\cite{ShenEiter16a}),
\item addresses the memory issues that plague some other implementations,
\item uses multi-processing to improve performance,
\item allows early termination when desired number of solutions found,
\item solves harder programs faster, and
\item permits solving programs on a typical laptop computer that were
previously beyond the capabilities of other solvers with reasonable resources.
\end{itemize}

For the future, we plan to enhance our solver further by allowing the use of
distributed processing to make it reasonable to solve more programs. We also
plan to look at generalizing certain heuristics that could significantly reduce
the search space for certain classes of programs. Finally, we plan to look at
techniques such as \emph{multi-shot solving}
\cite{GebserKaminskiKaufmannSchaub14a} to avoid the need for repeatedly
restarting the ASP solver from scratch.

We observe that with small modification our implementation could solve GI-log
programs \cite{ZhangWangZhang15a}, though a suitable front end is needed to
handle the program syntax.

Follow-up work will be to explore applications for the language as solver
improvements make its use more practical. Promising areas include planning \&
scheduling, policy management, diagnostics, and computer-assisted decision
making.

\section*{Acknowledgments}

We would like to acknowledge the efforts of Yi-Dong Shen and Thomas Eiter
whose work provided a positive influence on the development of our solver.
We thank Evgenii Balai for providing the front end to our solver by adding
an extra option to {\tt ELPS} to output the ASP translation. We couldn't
produce our results without the use of {\tt gringo} \& {\tt clingo} from the
Potassco team, especially Roland Kaminski and Ben Kaufmann. Finally, we
thank the anonymous reviewers for their comments and suggestions.

\bibliographystyle{splncs03}
\bibliography{ESbiblio}

\newpage
\appendix
\section*{Appendix: Proof of Semantic Equivalence to Shen-Eiter Semantics}

In this appendix, we prove the equivalence between the semantics of epistemic
logic programs (Section~\ref{SyntaxAndSemantics}) and the semantics of logic
programs with epistemic negation as defined by Shen \& Eiter in
\cite{ShenEiter16a}.

For the benefit of the reader, the following table summarizes the notational
differences between our language and that of Shen \& Eiter for semantically
equivalent forms.

\noindent
\begin{center}
\begin{tabular}{|c|c||l|}\hline\vspace{-0.01cm}
{\small\bf\textsc{Our Notation}}  ~&~
{\small\bf\textsc{SE Notation}}~~&~ {\small\bf\textsc{Terminology}} ~\\\hline\hline
$\neg$          & $\sim$          &~ classical (strong) negation    ~\\\hline
$\ASPnot$       & $\neg$          &~ default negation               ~\\\hline
$\leftASPnot\K$ & $\enaf$         &~ epistemic negation             ~\\\hline
$\K$            & $\neg\enaf$     &~ ~\\\hline
$\M$            & $\enaf\neg$     &~ ~\\\hline
$\leftASPnot\M$ & $\neg\enaf\neg$ &~ ~\\\hline
\end{tabular}
\end{center}

Let $\Pi$ be a ground epistemic logic program. Let $r$ be a rule in $\Pi$.
In this appendix, we use the following short hand to represent $r$:\vspace{-0.1cm}  
\begin{equation} \label{rule} 
H \leftarrow B^+, \ASPnot B^-, \k L_1, ~\mbox{\tt not}\k L_2, \m L_3, ~\mbox{\tt not}\m L_4\vspace{-0.1cm}
\end{equation} 
where $H$ is ~$h_1 \ASPor \ldots \ASPor h_n$~ and 
$B^+$,  $B^-$, $L_1, ..., L_4$ are sets of objective literals.
For $L = \{\ell_1,...,\ell_m\}$, $x\: L$ stands for $\{x\: \ell_1, ..., x\: \ell_m\}$
with $x \in \{\mbox{\tt not},\k,\leftASPnot\k,\m,\leftASPnot\m\}$.\\
\noindent
In Shen-Eiter notation \cite{ShenEiter16a}, the above rule will become the
following:\vspace{-0.1cm}
\begin{equation} \label{rulese} 
H \leftarrow B^+, ~\neg B^-, ~\neg \enaf L_1, \enaf L_2, \enaf \neg L_3, ~\neg  \enaf \neg L_4\vspace{-0.1cm}
\end{equation}

We denote by $SE(x)$ the corresponding syntactic form of $x$ in Shen-Eiter
notation; e.g., $SE(\Pi)$ denotes the program obtained by translating $\Pi$ to
Shen-Eiter notation. We further define $\Ep{SE(\Pi)}$ as being equivalent to
$SE(\Ep{\Pi})$ (see Definition \ref{EpDefinition}). When clear from the
context, we may use $\Phi_W$ to mean $SE(\Phi_W\hspace{-0.05cm})$, $r$ to
mean $SE(r)$, etc.

In proving the correspondence between the two semantics, we make the following 
assumptions and notes:

\begin{itemize}\vspace{-0.15cm}
\item The presence of classical negation can be eliminated using standard
transformation (e.g., the one proposed in \cite{GelfondLifschitz91a}). For this
reason, we will assume that programs in consideration do not contain classical
negation. 

\item Shen-Eiter programs allow $\enaf \neg \ell$\hspace{0.1cm}
(\hspace{0.01cm}$\mbox{\tt not}\hspace{0.1cm}\K\ASPnot\ell$ in our notation)
which we consider equivalent to $\m\ell$ in our syntax.

\item In Shen-Eiter programs, $\neg \neg \ell$\hspace{0.1cm}
(\hspace{0.01cm}$\mbox{\tt not}\ASPnot\ell\hspace{0.05cm}$ in our notation) is
treated as $\ell$.
\end{itemize}\vspace{-0.15cm}

We use ``$X$-reduct'' where $X$ is ``KLS'' (to refer to the modal reduct
discussed in this paper) or ``SE'' (to refer to the epistemic reduct proposed
by Shen \& Eiter in \cite{ShenEiter16a}). For ASP programs with nested default
negation, a ``nested-negation removal transformation'' refers to the usual
substitution of $\mbox{\tt not}\ASPnot \ell\ $ with $\ASPnot \ell^\prime\ $
plus the addition of rule $\ell^\prime \leftarrow \mbox{\tt not}~\ell$ where
$\ell^\prime$ is a fresh literal and answer sets are modulo $\ell^\prime$ as
described in \cite{LifschitzTangTurner99a}. 
We use ``Gelfond-Lifschitz transformation'' to refer to the reduction from
logic programs with default negation to positive programs as per
\cite{GelfondLifschitz91a}.
We say that a rule whose body contains a conjunct of the form $\neg \top$ in
Shen-Eiter notation is a {\em useless} rule as this rule is always satisfied
(body is false) and cannot be used to justify anything.  

In the following, whenever we refer to a rule $r$, we mean a rule with
sets of atoms $L_1, \ldots, L_4$ as in~\eqref{rule}. We use $\Pi$ to denote an
arbitrary but fixed ground epistemic logic program. Furthermore, for a program
$\Pi,$ a set of sets of atoms $W,$ a set $A \in W,$ and a rule $r \in \Pi,$ by
the \emph{reduct of $r$ in $(\Pi^W)^A$} (resp. \emph{reduct of $SE(r)$ in
$(SE(\Pi)^{\Phi_W})^A$)} we mean the rule (if it exists) obtained from $r$
after:
\begin{enumerate}
\item computing the modal reduct of $\Pi$ with respect to $W$ according to
Definition~\ref{ModRedDef} (resp. the epistemic reduct of $SE(\Pi)$ with respect
to $\Phi_W$ as per \cite{ShenEiter16a}); and then
\item computing the Gelfond-Lifschitz transformation on the result with respect
to $A.$
\end{enumerate}

\begin{lemma}\label{l01}
Let $r$ be a rule in $\Pi$, $W$ be a collection of sets of atoms in $\Pi$, and
$A \in W.$  If there exists some $\ell \in L_1$ of $r$ such that
$W \not\models \k \ell$ then there is no reduct of $r$ in $(\Pi^W)^A$ and
either there is no reduct of $r$ in $(SE(\Pi)^{\Phi_W})^A$ or the reduct of $r$
is a useless rule.
\end{lemma} 
\noindent 
{\bf Proof.} 
$W \not\models \k \ell$ implies that $r$ is removed in the KLS-reduct and hence
$(\Pi^W)^A$ does not contain the reduct of $r$. 

$W \not\models \k \ell$ means that $W \models \ASPnot \k \ell$. 
This means that $\enaf\ \ell \in \Phi_W$. Thus, the SE-reduct of $r$ in
$SE(\Pi)$ is a rule whose body contains $\neg \top$, a useless rule. This
implies that $(SE(\Pi)^{\Phi_W})^A$ contains no reduct of $r$ if the
Gelfond-Lifschitz transformation removes it; or a useless rule, which is the
reduct of $r$.
\hfill $\Box$

\medskip
\begin{lemma}\label{l02} 
Let $r$ be a rule in $\Pi,$ $W$ be a collection of sets of atoms in $\Pi,$ and
$A \in W.$ If there exists some $\ell \in L_2 \cap A$ such that
$W \not\models \ASPnot \k \ell$, then $(\Pi^W)^A$ and $(SE(\Pi)^{\Phi_W})^A$
contain no reduct of $r$.
\end{lemma} 
\noindent
{\bf Proof.} 
$W \not\models \ASPnot \k \ell$ implies that there exists some $S \in W$ such
that $\ell \not\in S$. This implies that $W \not\models \enaf \ell$ and so
$\enaf \ell \not\in \Phi_W$.

Let $r'$ and $r''$ be the result obtained by applying the KLS-reduct and the
SE-reduct on $r$, respectively.
We have $r' \in \Pi^W$ and $r'' \in SE(\Pi)^{\Phi_W}$.

Since $W \not\models \ASPnot \k \ell$, $\ASPnot \ell$ occurs in the body of
$r'$. Because $\ell \in A$, the Gelfond-Lifschitz transformation will remove
the rule $r' $ from $\Pi^W$ when constructing $(\Pi^W)^A$, i.e., $(\Pi^W)^A$
does not contain the reduct of $r$. 

Similarly, since $\enaf \ell \not\in \Phi_W$, $\neg \ell$ occurs in the body 
 of $r''$ in $SE(\Pi)$, and $\ell \in A$, the Gelfond-Lifschitz transformation
will remove the rule $r'' $ when constructing $(SE(\Pi)^{\Phi_W})^A$, i.e.,
$(SE(\Pi)^{\Phi_W})^A$ does not contain the reduct of $r$.

So, in this case, neither $(\Pi^W)^A$ nor $(SE(\Pi)^{\Phi_W})^A$ contains the
reduct of $r$. 
\hfill $\Box$

\medskip
\begin{lemma}\label{l03} 
Let $r$ be a rule in $\Pi$, $W$ be a collection of sets of atoms in $\Pi$ and
$A \in W$. If there exists some $\ell\in L_3$ such that $W \not\models \m \ell$
then $(\Pi^W)^A$ and $(SE(\Pi)^{\Phi_W})^A$ contain no reduct of $r$ or a
useless reduct of $r$ with respect to $A$. 
\end{lemma} 
\noindent 
{\bf Proof.} 
$W \not\models \m \ell$ implies that there exists no $S \in W$ such that 
$\ell \in S$, i.e., for every $S \in W$, $\ell \not\in S$. This implies that
$W \not\models \enaf \neg \ell$ and so $\enaf \neg \ell \not\in \Phi_W$. 
This also implies that $\ell \not\in A$. 

\newpage

In this case, we observe that $\ASPnot\leftASPnot\ell$ occurs in the body of
the KLS-reduct $r'$ of $r$ in $\Pi^W$ and $\neg \neg \ell$ occurs in the body
of the SE-reduct $r''$ of $r$ in $SE(\Pi)^{\Phi_W}\hspace{-0.07cm}.$ Shen
\& Eiter treat $\neg \neg \ell$ as $\ell$. Since $\ell \not \in A$,
$(SE(\Pi)^{\Phi_W})^A$ will contain no reduct of $r$ (if it is removed by the
Gelfond-Lifschitz reduct) or a useless rule with respect to $A$. 

In KLS, nested default negation is treated differently. The nested-negation
removal transformation results in $\ASPnot not\_\ell$ appearing in the body of
$r'$. Furthermore, $\Pi^W$ contains the rule
``$not\_\ell \leftarrow \ASPnot \ell$''
where $not\_\ell$ is a fresh atom representing $\ASPnot \ell$.

Let $\Pi^n$ be the
programs obtained from $\Pi^W$ after the nested-negation removal
transformation. Because $\ell \not\in A$, any answer set $A'$ of $\Pi^n$ whose
reduction results into $A$ must contain $not\_\ell$. This implies that the
Gelfond-Lifschitz transformation will remove $r' $ from $\Pi^n$  when
constructing $(\Pi^W)^A$.

So, in this case, $(\Pi^W)^A$ does not contain the reduct of $r$ and 
$(SE(\Pi)^{\Phi_W})^A$ contain no reduct of $r$ or a useless reduct of 
$r$ with respect to $A$. 
\hfill $\Box$

\medskip
\begin{lemma}\label{l04} 
Let $r$ be a rule in $\Pi$, $W$ be a collection of sets of atoms in $\Pi$ and
$A \in W$. If there exists some $\ell \in L_4$ such that
$W \not\models \ASPnot \m \ell$ then $(\Pi^W)^A$ contains no reduct of $r$
and $(SE(\Pi)^{\Phi_W})^A$ contains no reduct of $r$ or a {\em useless} reduct
of $r$.
\end{lemma} 
\noindent 
{\bf Proof.} 
$W \not\models \ASPnot \m \ell$ implies that $r$ is removed in the KLS-reduct
and hence $(\Pi^W)^A$ does not contain the reduct of $r$. 

$W \not\models \ASPnot \m \ell$ implies that there exists some $S \in W$ such
that $\ell \in S$. It means that $W \not\models \k \ASPnot \ell$ and hence
$\enaf \neg \ell \in \Phi_W$. Thus, the SE-reduct of $r$ in $SE(\Pi)^{\Phi_W}$
is a rule whose body contains $\neg \top$. Again, this means that
$(SE(\Pi)^{\Phi_W})^A$ contains no reduct of $r$ or a useless rule which is the
reduct of $r$.
\hfill $\Box$

\medskip 
\noindent 
\begin{theorem}\label{th1}
Let $W$ be a collection of sets of atoms in $\Pi$. 
 $W$ is a world view of $\Pi$ under the KLS semantics if and only if 
$W$ is a candidate world view of $SE(\Pi)$ w.r.t. $\Phi_W$.
\end{theorem} 
\noindent{\bf Proof.} 
Observe that for each $\enaf \ell$ occurring in $SE(\Pi)$, it holds that  
\begin{itemize} 
\item if $\enaf \ell \in \Phi_W$ then  $\enaf \ell$ is true in $W$; or 
\item if $\enaf \ell \in \Ep{SE(\Pi)} \setminus \Phi_W$ then $\enaf \ell$ is
false in $W$.
\end{itemize} 

Let $L_2^t = \{\ell \mid \ell \in L_2 \land W \models \ASPnot \k \ell\}$ and
$L_2^f = \{\ell \mid \ell \in L_2 \land W  \not\models \ASPnot \k \ell\}$. 
Let $L_3^t = \{\ell \mid \ell \in L_3 \land W \models \m \ell\}$ and
$L_3^f = \{\ell \mid \ell \in L_3 \land W \not \models \m \ell\}$.
Consider $A \in W$ and let $r$ be a rule of the form~\eqref{rule}. 
Lemmas~\ref{l01}-\ref{l04} show that if $(\Pi^W)^A$ or $(SE(\Pi)^{\Phi_W})^A$
contains a useful reduct of $r$ then the following conditions hold: 
\begin{itemize}
\item $L_1 \subseteq A$ ~(Lemma~\ref{l01}). 
\item for every $\ell \in L_2^f$, $\ell \not\in A$ ~(Lemma~\ref{l02}). 
\item $L_3^f = \varnothing$ ~(Lemma~\ref{l03}). 
\item $L_4 \cap A = \varnothing$ ~(Lemma~\ref{l04}). 
\item $B^- \cap A = \varnothing$ ~(otherwise, the rule will be removed by the
Gelfond-Liftschitz transformation).
\end{itemize} 

Under the above conditions, the reduct of $r$ in $\Pi^W$ is 
\[
r': \:\:\:\:H \leftarrow B^+, \ASPnot B^-,    L_1, \ASPnot L_2^f,   \ASPnot  L_4 
\]
and its reduct in $(SE(\Pi))^{\Phi_W}$ is 
\[
r'': \:\:\:\: H \leftarrow B^+, \ASPnot B^-,   \neg \neg L_1, \neg L_2^f,   \neg \neg \neg L_4 
\]
Since $\neg \neg L = L$ as stated in Shen \& Eiter's proposal and because
$L_4 \cap A = \varnothing$ and $L^f_2 \cap A = \varnothing$, we can conclude that
the reduct of $r$ in $(\Pi^W)^A$ and in  $(SE(\Pi)^{\Phi_W})^A$ is 
\[
r^+: \:\:\:\: \ell_1 \ASPor \ell_2 \ldots \ASPor \ell_n \leftarrow B^+, L_1 
\]
In other words, we have that  $(\Pi^W)^A$ is identical\footnote{
  with the exception of the useless rules in different programs 
  and some extra fresh atoms such as ~$not\_a$~ representing $\ASPnot a$~
  and the rule ~$not\_a \leftarrow \leftASPnot a$~ introduced to deal with
  $\leftASPnot\leftASPnot a$; such atoms can be eliminated by using the
  splitting set theorem and the fact that, by construction,
  $\ASPnot\leftASPnot a$~ appears in ~$\Pi^W$ only if ~$a \not \in A$
} to $(SE(\Pi)^{\Phi_W})^A$. As such, if $W$ is a world view of $\Pi$, $A$ is
an answer set of $(\Pi^W)^A$, i.e., $A$ is also an answer set of
$(SE(\Pi)^{\Phi_W})^A$. Conversely, 
if $W$ is a candidate world view of $\Pi$ with respect to $\Phi_W$ then 
$A$ is an answer set of $(SE(\Pi)^{\Phi_W})^A$ and hence 
 also an answer set of $(\Pi^W)^A$. Since this holds for every $A \in W$, 
 we have the conclusion of the theorem.  
  \hfill $\Box$ 

\end{document}